\documentclass{article}

\usepackage{arxiv}

\usepackage[utf8]{inputenc} 
\usepackage[T1]{fontenc}    
\usepackage{hyperref}       
\usepackage{url}            
\usepackage{booktabs}       
\usepackage{amsfonts}       
\usepackage{nicefrac}       
\usepackage{microtype}      
\usepackage{lipsum}

\usepackage{tikz}
\usetikzlibrary{positioning}
\usetikzlibrary{shadows}
\usetikzlibrary{shapes}
\usetikzlibrary{calc}
\usetikzlibrary{fit}
\usetikzlibrary{arrows}
\usepackage{todonotes}

\newenvironment{concept}{%
\begin{tikzpicture}[
	every shadow/.style={opacity=.2},
	files/.style={
		draw,
		double copy shadow={left color=gray!20, right color=gray!50, middle color=white},
		left color=gray!10,
		right color=gray!30,
		middle color=white,
	},
	execution/.style={
		draw,
		thick,
		rounded corners,
		inner sep=1ex,
	},
	output/.style={
		draw,
		thick,
		rounded corners,
		drop shadow,
		left color=red!10,
		right color=red!25,
		middle color=white,
		inner sep=1ex,
	},
	file/.style={
		draw,
		tape,
		tape bend top=none,
		tape bend height=1mm,
		left color=teal!10,
		right color=teal!30,
		middle color=white,
		drop shadow,
	},
	rect/.style={
		draw,
		rectangle,
		left color=gray!10,
		right color=gray!30,
		middle color=white,
	},
	node/.style={
		draw,
		circle,
		text width=4pt,
		minimum size=9pt,
		inner sep=1pt,
		font=\tiny,
	},
	font=\footnotesize,
	node distance=4mm,
	arrows=->,
	>=stealth,
]}{\end{tikzpicture}}

\title{ASPARTIX-V21}

\author{
Wolfgang Dvo{\v{r}}{\'{a}}k\\
  Institute of Logic and Computation\\
  TU Wien\\
  Vienna, Austria\\
  \texttt{dvorak@dbai.tuwien.ac.at} \\
   \And
  Matthias K\"{o}nig\\
  Institute of Logic and Computation\\
  TU Wien\\
  Vienna, Austria\\
  \texttt{mkoenig@dbai.tuwien.ac.at} \\
   \And
 Johannes P. Wallner \\
  Institute of Software Technology\\
  Graz University of Technology\\
  Graz, Austria \\
  \texttt{wallner@ist.tugraz.at} \\
  \And
  Stefan Woltran\\
  Institute of Logic and Computation\\
  TU Wien\\
  Vienna, Austria\\
  \texttt{woltran@dbai.tuwien.ac.at} \\
}

\begin{document}
\maketitle

\begin{abstract}
In this solver description we present ASPARTIX-V, in its 2021 edition, which participates in the International Competition on Computational Models of Argumentation (ICCMA) 2021. 
\mbox{ASPARTIX-V} is capable of solving all classical (static) reasoning tasks part of ICCMA'21 and extends the ASPARTIX system suite by incorporation of 
recent ASP language constructs (e.g.\ conditional literals), 
domain heuristics within ASP, and 
multi-shot methods. 
In this light ASPARTIX-V deviates from the traditional focus of ASPARTIX on monolithic approaches (i.e., one-shot solving via a single ASP encoding) to further enhance performance.
\end{abstract}

\section{Solver Description}
\begin{figure}[h]
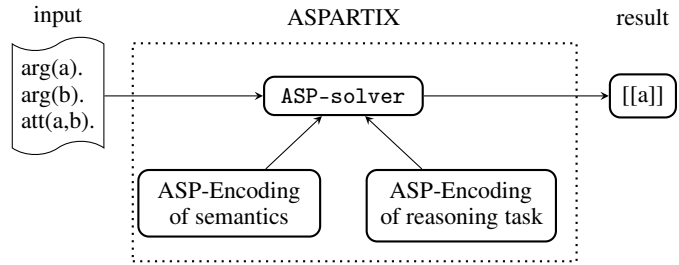

\centering
\begin{concept}
\node [execution, minimum width=60pt] (iter) {\texttt{ASP-solver}}; 
\node [node distance=33pt,draw,tape, left= 60pt of iter, align=left] (instance) {
arg(a).\\
arg(b).\\
att(a,b).
};
\node [node distance=30pt,above of=instance] {input};
\node [node distance=25pt,execution, below left=20pt and -22pt of iter, minimum width=18mm] (semantics) {
\begin{minipage}{21mm}
\centering
ASP-Encoding of semantics
\end{minipage}
};
\node [node distance=25pt,execution, below right= 20pt and -22pt of iter, minimum width=18mm] (task) {
\begin{minipage}{22mm}
\centering
ASP-Encoding of reasoning task
\end{minipage}
};

\node [node distance=33pt,execution, right= 70pt of iter, align=center] (postproc) {[[a]]};
\node [node distance=30pt,above of=postproc] {result};8
\node [node distance=30pt,above of=iter] {ASPARTIX};

\draw[dotted,thick] (-2.8,-2.2) rectangle (3.10,0.7);

\draw (instance) -> (iter);
\draw (semantics) to (iter);
\draw (task) to (iter);
\draw (iter) -> (postproc);
\end{concept}
\caption{Basic workflow of ASPARTIX-V}
\label{fig:iterative_workflow}
\end{figure}

In this paper we describe ASPARTIX-V (Answer Set Programming Argumentation Reasoning Tool - Vienna) in its 2021 edition. 
ASPARTIX-V21 solves reasoning tasks on argumentation frameworks (AFs)~\cite{Dung95} and is an update of
ASPARTIX-V19 \cite{DvorakRWW20a} which itself is based on earlier versions of ASPARTIX and its derivatives~\cite{EglyGW10,DvorakGWW2013,DvorakGLW14,GagglMRWW2015,aspartixv}. 
Given an AF as input, in the format of $\mathtt{apx}$, ASPARTIX-V delegates the main reasoning to an answer set programming (ASP) solver~(e.g.~\cite{GebserKKS14}), with answer set programs encoding the argumentation semantics and reasoning tasks. 
The basic workflow is illustrated in Figure~\ref{fig:iterative_workflow}, i.e., the AF is given in the $\mathtt{apx}$ format (facts in the ASP language), and the AF semantics and reasoning tasks are encoded via ASP rules, possibly utilizing further ASP language constructs. 
In the next section we highlight specifics of the current version and in particular differences to prior versions. 
ASPARTIX, and its derivatives, are available online under
\begin{center}
\url{https://www.dbai.tuwien.ac.at/research/argumentation/aspartix/}
\end{center}

\section{The ASPARTIX System Suite and its ASPARTIX-V Editions}
The ASPARTIX system \cite{EglyGW10,DvorakGRWW20b} suite exploits answer-set programming to implement efficient reasoning for several abstract argumentation formalisms. The core of the ASPARTIX system implements a broad collection of semantics for abstract argumentation frameworks and several enhancements and generalizations of AFs (e.g., by preferences or recursive attacks) have been 
incorporated to the ASPARTIX system suite.
The ASPARTIX-V editions~\cite{aspartixv,DvorakRWW20a} of ASPARTIX implement an ICCMA compatible interface~\footnote{\url{http://argumentationcompetition.org/2021/SolverRequirements.pdf}} for reasoning with AFs together with several optimizations towards the participation in ICCMA.

ASPARTIX-V21 is essentially an update of the preceding ASPARTIX-V19 version. 
Both versions deviate from classical ASPARTIX design virtues in order to improve the performance.
First, while traditional ASPARTIX encodings are modular in the sense 
that fixed encodings for semantics can be combined with the generic encodings of reasoning tasks, 
we use semantics encodings specific to a reasoning task.
Second, when appropriate we apply multi-shot methods for reasoning which is in contrast to the earlier focus on so-called monolithic encodings, where one uses a single ASP-encoding and runs the solver only once (as illustrated in Figure~\ref{fig:iterative_workflow}).
Third, we make use of advanced features of the ASP-language, and utilize the ASP-solvers clingo v5.3.0 and v4.4.0~\cite{GebserKKS14}.\footnote{\url{https://potassco.org/}}

The updates according to the ASPARTIX-V19 version are the following.
ASPARTIX-V21 is the first version implementing the counting extensions tasks, as required by ICCMA'21. 
Moreover, the shell scripts implementing the ICCMA interface have been rewritten to avoid issues with concurrent calls to the systems.

\section{Supported Reasoning Tasks}
ASPARTIX-V21 supports all the standard tasks of ICCMA 2021 and the enumerating extensions task from previous editions of ICCMA. However, it does not support the dynamic settings of the special track.
That is, ASPARTIX-V21 supports complete (CO), preferred (PR), stable (ST), semi-stable (SST), stage (STG), and 
ideal (ID) semantics.
For each of the semantics it supports the following reasoning tasks.
\begin{itemize}
 \item Some Extension (SE): Given an AF, determine some extension.
 \item Enumerating Extensions (EE): Given an AF, enumerate all extensions.
 \item Counting Extensions (CE): Given an AF, determine the number of its extensions.
 \item Decide Credulous Acceptance (DC): Given an AF and some argument, decide whether the given argument is credulously inferred.
 \item Decide Skeptical Acceptance (DS): Given an AF and some argument, decide whether the given argument is skeptically inferred.
\end{itemize}
The source code of of the competition version ASPARTIX-V21 is available at the following link:

\begin{center}
  \url{https://www.dbai.tuwien.ac.at/research/argumentation/aspartix/dung.html#competition}
\end{center}

\section{Implementation Details}

When not stated otherwise we follow the standard ASPARTIX approach, i.e.\
for a supported semantics we provide an ASP-encoding such that 
when combined with an AF in the $\mathtt{apx}$ format the answer-sets of the program 
are in a one-to-one correspondence with the extensions of the AF.
Given an answer-set of such an encoding the corresponding extension is given by the  \texttt{in($\cdot$)} predicate, i.e., an
argument \texttt{a} is in the extensions iff  \texttt{in(a)} is in the answer-set.
With such an encoding we can exploit a standard ASP-solver to:
compute some extension (SE) by computing an answer-set;
enumerate all extensions (EE) by enumerating all answer-sets;
count extensions (CE) by counting the answer-sets;
decide credulous acceptance (DC) of an argument $a$ by adding the constraint \texttt{:- not in(a)} to the program
and testing whether the program is satisfiable,
i.e., $a$ is credulously accepted if there is at least one answer set; and
decide skeptical acceptance (DS) of an argument $a$ by adding the constraint \texttt{:- in(a)}  to the program
and testing whether the program is unsatisfiable,
i.e., $a$ is skeptically accepted if there is no answer set.

For the implementation of certain semantics and reasoning tasks we deviate from the above described standard way of ASPARTIX. 
In the following we briefly describe these modifications:

For credulous and skeptical semantics with complete, preferred, 
and ideal semantics 
we do not need to consider the whole argumentation framework but only those arguments that have a directed path to the query argument
(notice that this does not hold true for stable, semi-stable and stage semantics). 
That is, we perform pre-processing on the given AF that removes arguments without a directed path to the queried argument before starting the reasoning with ASP-solver. We do so for the reasoning tasks DC-CO (DC-PR), DC-ID, and DS-PR.

For computing the ideal extension (SE-ID, EE-ID) we follow a two-shot strategy.
That is, we first use an encoding for complete semantics and the brave reasoning mode of clingo to compute all arguments
that are credulously accepted/attacked w.r.t.\ preferred semantics. 
Second, we use the outcome of the first call together with an encoding that computes a fixed-point which corresponds to the ideal extension.
For reasoning with ideal semantics (DC-ID, DS-ID) we use an encoding for ideal sets 
and perform credulous reasoning on ideal sets as described in the first paragraph of this section.

Clingo provides an option to add user-specific domain heuristics to the ASP program
which in particular allow to select the answer-sets that are subset-maximal/minimal w.r.t.\ a specified predicate.
We use such heuristics for preferred semantics (EE-PR, SE-PR, CE-PR) by using an encoding for complete semantics and then identifying the subset-maximal answer-sets w.r.t.\ the \texttt{in($\cdot$)} predicate.
Moreover, we use domain heuristics for semi-stable and stage semantics.
Here we start from an encoding for complete extensions, conflict-free sets respectively,
and use domain heuristics to compute the subset-maximal ranges\footnote{The range of a set of arguments $S$ is the set of arguments that are either contained in $S$ or attacked by an argument in $S$.}
a complete extension, a conflict-free set respectively, can have.
The answer-sets corresponding to these ranges are then exploited for computing some semi-stable (SE-SST) or stage extension (SE-STG).
However, the domain heuristics only return one witnessing answer-set for each maximal range and thus this technique is not directly applicable to the corresponding enumeration and counting tasks (we would miss certain extensions if several extensions have the same range).

Semi-stable extensions correspond to those complete labellings for which
the set of undecided arguments is subset-minimal. 
For enumerating or counting semi-stable extensions (EE-SST, CE-SST), multiple answer-sets 
possessing the same subset-minimal set of undecided arguments can exist.
In our approach, we utilize an encoding for complete semantics 
extended by an \texttt{undec($\cdot$)} predicate 
and process the answer-sets. We check whether
models without \texttt{undec($\cdot$)} predicate have been computed;
in that case, semi-stable extensions coincide with stable extensions.
In the other case,
we compute all subset-minimal sets among all undecided sets
using the set class in python and return 
the corresponding models.

For enumerating and counting stage extensions (EE-STG, CE-STG) we use a multi-shot strategy. 
First, we use domain heuristic to compute the maximal ranges w.r.t.\ naive semantics\footnote{Naive sets are subset-maximal conflict-free sets and as each range maximal conflict-free set is also subset-maximal it is sufficient to only consider naive sets.}.
Second, for each of the maximal ranges we start another ASP-encoding which computes conflict-free sets with 
exactly that range (this is equivalent to computing stable extension of a restricted framework).
Each of these extensions corresponds to a different stage extension of the AF.

For reasoning with semi-stable and stage semantics (DC-SST, DS-SST, DC-STG, DS-STG) we use a multi-shot strategy similar to that for enumerating the stage extensions.
First we use domain heuristics to compute the maximal ranges w.r.t.\ complete and naive semantics.
In the second step we iterate over these ranges and perform skeptical, credulous respectively, reasoning over
complete extensions, conflict-free sets respectively, with the given range.
For skeptical acceptance, we answer negatively as soon as a counterexample to a positive answer is found when iterating the extensions; otherwise, after 
processing all maximal ranges we answer with YES. 
Analogously, for credulous acceptance,
we check in each iteration whether we can report a positive answer; otherwise, after processing all maximal ranges, we return NO.

\section*{Acknowledgements}
This work was supported by the
Vienna Science and Technology Fund (WWTF) through project ICT19-065,
and the Austrian Science Fund (FWF) through projects P30168, P32830, and Y698.

The authors are grateful to all the people that contributed to the ASPARTIX project over the years and in particular to Anna Rapberger for her contributions to the preceding version ASPARTIX-V19.
\newpage
\bibliographystyle{unsrt}  
\bibliography{references} 

\begin{thebibliography}{1}

\bibitem{Dung95}
Phan~Minh Dung.
\newblock {On the Acceptability of Arguments and its Fundamental Role in
  Nonmonotonic Reasoning, Logic Programming and n-Person Games}.
\newblock {\em Artif. Intell.}, 77(2):321--358, 1995.

\bibitem{DvorakRWW20a}
Wolfgang Dvo\v{r}{\'{a}}k, Anna Rapberger, Johannes~Peter Wallner, and Stefan
  Woltran.
\newblock {ASPARTIX-V19} - an answer-set programming based system for abstract
  argumentation.
\newblock In Andreas Herzig and Juha Kontinen, editors, {\em Foundations of
  Information and Knowledge Systems - 11th International Symposium, FoIKS 2020,
  Dortmund, Germany, February 17-21, 2020, Proceedings}, volume 12012 of {\em
  Lecture Notes in Computer Science}, pages 79--89. Springer, 2020.

\bibitem{EglyGW10}
Uwe Egly, Sarah~Alice Gaggl, and Stefan Woltran.
\newblock Answer-set programming encodings for argumentation frameworks.
\newblock {\em Argument {\&} Computation}, 1(2):147--177, 2010.

\bibitem{DvorakGWW2013}
Wolfgang Dvo\v{r}\'{a}k, Sarah~A. Gaggl, Johannes~P. Wallner, and Stefan
  Woltran.
\newblock Making use of advances in answer-set programming for abstract
  argumentation systems.
\newblock In Hans Tompits, Salvador Abreu, Johannes Oetsch, J\"{o}rg
  P\"{u}hrer, Dietmar Seipel, Masanobu Umeda, and Armin Wolf, editors, {\em
  Proc.~INAP, Revised Selected Papers}, volume 7773 of {\em Lecture Notes in
  Artificial Intelligence}, pages 114--133. Springer, 2013.

\bibitem{DvorakGLW14}
Wolfgang Dvo\v{r}\'{a}k, Sarah~Alice Gaggl, Thomas Linsbichler, and
  Johannes~Peter Wallner.
\newblock Reduction-based approaches to implement {M}odgil's extended
  argumentation frameworks.
\newblock In Thomas Eiter, Hannes Strass, Miroslaw Truszczynski, and Stefan
  Woltran, editors, {\em Advances in Knowledge Representation, Logic
  Programming, and Abstract Argumentation - Essays Dedicated to Gerhard Brewka
  on the Occasion of His 60th Birthday}, volume 9060 of {\em Lecture Notes in
  Computer Science}, pages 249--264. Springer, 2015.

\bibitem{GagglMRWW2015}
Sarah~Alice Gaggl, Norbert Manthey, Alessandro Ronca, Johannes~Peter Wallner,
  and Stefan Woltran.
\newblock Improved answer-set programming encodings for abstract argumentation.
\newblock {\em Theory and Practice of Logic Programming}, 15(4-5):434--448,
  2015.

\bibitem{aspartixv}
Alessandro Ronca, Johannes~Peter Wallner, and Stefan Woltran.
\newblock {ASPARTIX-V:} utilizing improved {ASP} encodings.
\newblock \url{http://argumentationcompetition.org/2015/pdf/paper_11.pdf},
  2015.

\bibitem{GebserKKS14}
Martin Gebser, Roland Kaminski, Benjamin Kaufmann, and Torsten Schaub.
\newblock Clingo = {ASP} + control: Preliminary report.
\newblock {\em CoRR}, abs/1405.3694, 2014.

\bibitem{DvorakGRWW20b}
Wolfgang Dvo\v{r}{\'{a}}k, Sarah~Alice Gaggl, Anna Rapberger, Johannes~Peter
  Wallner, and Stefan Woltran.
\newblock The {ASPARTIX} system suite.
\newblock In Henry Prakken, Stefano Bistarelli, Francesco Santini, and Carlo
  Taticchi, editors, {\em Computational Models of Argument - Proceedings of
  {COMMA} 2020, Perugia, Italy, September 4-11, 2020}, volume 326 of {\em
  Frontiers in Artificial Intelligence and Applications}, pages 461--462. {IOS}
  Press, 2020.

\end{thebibliography}
\end{document}